\documentclass[conference]{IEEEtran}
\IEEEoverridecommandlockouts

\usepackage{cite}
\usepackage{amsmath,amssymb,amsfonts}
\usepackage{algorithmic}
\usepackage{graphicx}
\usepackage{textcomp}
\usepackage{xcolor}
\def\BibTeX{{\rm B\kern-.05em{\sc i\kern-.025em b}\kern-.08em
    T\kern-.1667em\lower.7ex\hbox{E}\kern-.125emX}}

\usepackage{url}
\usepackage{doi}
\usepackage{amsmath}
\usepackage{booktabs}

\begin{document}

\title{\fontsize{20pt}{22pt}\selectfont HAVIR: HierArchical Vision to Image Reconstruction using CLIP-Guided Versatile Diffusion}

\author{%
\IEEEauthorblockN{Shiyi Zhang}
\IEEEauthorblockA{Department of Biomedical Engineering,\\
Southern University of Science and Technology,\\ Shenzhen, China\\
Email: sy.zhang4@siat.ac.cn}
\and
\IEEEauthorblockN{Dong Liang}
\IEEEauthorblockA{Research Center for Medical AI,\\
Shenzhen Institutes of Advanced Technology,\\
Chinese Academy of Sciences, Shenzhen, China\\
Email: dong.liang@siat.ac.cn}
\and
\IEEEauthorblockN{Hairong Zheng}
\IEEEauthorblockA{Research Center for Biomedical Imaging,\\
Shenzhen Institutes of Advanced Technology,\\
Chinese Academy of Sciences, Shenzhen, China\\
Email: hr.zheng@siat.ac.cn}
\and
\IEEEauthorblockN{Yihang Zhou\IEEEauthorrefmark{1}}
\IEEEauthorblockA{Research Center for Medical AI,\\
Shenzhen Institutes of Advanced Technology,\\
Chinese Academy of Sciences, Shenzhen, China\\
Email: yh.zhou2@siat.ac.cn}
\thanks{\IEEEauthorrefmark{1}Corresponding author.}
}

\maketitle

\begin{abstract}
The reconstruction of visual information from brain activity fosters interdisciplinary integration between neuroscience and computer vision. However, existing methods still face challenges in accurately recovering highly complex visual stimuli. This difficulty stems from the characteristics of natural scenes: low-level features exhibit heterogeneity, while high-level features show semantic entanglement due to contextual overlaps. 

Inspired by the hierarchical representation theory of the visual cortex, we propose the HAVIR model, which separates the visual cortex into two hierarchical regions and extracts distinct features from each. Specifically, the Structural Generator extracts structural information from spatial processing voxels and converts it into latent diffusion priors, while the Semantic Extractor converts semantic processing voxels into CLIP embeddings. These components are integrated via the Versatile Diffusion model to synthesize the final image. Experimental results demonstrate that HAVIR enhances both the structural and semantic quality of reconstructions, even in complex scenes, and outperforms existing models.
\end{abstract}


\section{Introduction}
Human brain processes complex visual information with remarkable efficiency, establishing multi level neural representations throughout the visual cortex\cite{brain1,brain2,brain3,retina}. By capturing neural responses to visual stimuli using functional magnetic resonance imaging (fMRI) and then leveraging generative artificial intelligence to reconstruct images, this approach provides novel possibilities for developing human-computer interactive system\cite{1-13}. Early approaches mostly relied on Variational Autoencoder (VAE) or generative adversarial networks (GANs) to synthesis images\cite{2-7,3-8,4-9,5-10,6-11,7-12}. Although these methods partially recovered the fundamental structure of the visual stimulus, they were unable to decode the rich semantic information inherent in neural activity, leading to a lack of perceptual detail and semantic accuracy. 

Motivated by the introduction of the large scale NSD dataset\cite{8-14}, Takagi et al.\cite{9-1} were the first to reconstruct high resolution images from NSD dataset using a latent diffusion model. Later studies such as Ozcelik et al.\cite{10-2}, UniBrain\cite{11-3-23}, MindEye\cite{12-4} and Mind-Diffuser\cite{14-6} used conditional diffusion models to incorporate semantic information into visual decoding, improving the representation of high-level features. Then, MindBridge\cite{13-5}, NeuroPictor\cite{Neuropictor}, Psychometry\cite{psychometry}, and BrainGuard\cite{BrainGuard} each designed a cross-subject decoding model, leading to notable progress.

However, existing methods perform inadequately in reconstruction tasks involving complex visual stimuli such as highly cluttered backgrounds, partially occluded or small-sized objects, and dense spatial structures. This limitation primarily arises from the characteristics of complex scenes: low-level features exhibit high heterogeneity (e.g., unstructured textures and occluded edges), while high-level features demonstrate semantic entanglement (e.g., multifaceted semantic information and ambiguous contextual relationships). Current methods struggle to achieve synergistic optimization that ensures faithful preservation of structural features while accurately representing the core semantic information.

To address these challenges, we propose a model named HAVIR, inspired by the hierarchical visual processing mechanisms of the brain\cite{brain3,retina}. On the one hand, neural activities in early visual areas respond strongly to low-level image features, such as shapes and positions\cite{nips21-1,nips21-2,nips21-3}. On the other hand, anterior visual areas primarily process high-level information, and their neural responses are highly correlated with the semantic content of the visual stimuli\cite{11-8,nips21-4}. These high-level features are more categorical and abstract than low-level ones in the image reconstruction. 

Following this principle, HAVIR reconstructs images from fMRI activity via complementary pathways: the Structural Generator predicts object layouts from spatial-processing voxels, while the Semantic Extractor retrieves semantic content from semantic-processing voxels. The pre-trained Versatile Diffusion\cite{mdb46} then integrates these low-level and high-level features to synthesize the final image. Experiments show HAVIR outperforms state-of-the-art methods in image reconstruction. Our contributions are summarized as follows:
\begin{enumerate}
\item We propose decomposing fMRI signals into structural and semantic voxels, reflecting the visual system's hierarchical processing. By extracting features separately for each type, this ``divide and conquer" strategy enhances visual information decoding while disentangling structure from semantics.
\item We design a Structural Generator and Semantic Extractor that decode spatial and semantic information from low-level and high-level visual cortex regions, respectively. These components transform fMRI signals into latent diffusion variables and CLIP semantic embeddings.
\item Considering neuroanatomical variability, we employ individualized brain region masks with manually delineated ROI boundaries rather than the standardized templates typically used in previous research. This enables precise brain decoding for specific subjects.
\end{enumerate}

\section{Related Works}
\subsection{Diffusion Models}
Diffusion Models (DMs)\cite{3-7,mdb8,Mdb13,14-6,4-9,11-3-23} are probabilistic generative models that generate samples through iterative denoising of Gaussian noise. To address computational limitations in pixel space processing, Latent Diffusion Models (LDMs) were developed to operate in compressed latent spaces. The framework utilizes a pretrained autoencoder that compresses images into low dimensional latent representations, significantly reducing computational costs while maintaining generation fidelity.

Expanding on LDM, the Versatile Diffusion (VD)\cite{mdb46} introduces a dual guidance mechanism that integrates CLIP features from images, text, or both, with a distinctive reverse process initialization strategy. VD is trained on datasets such as Laion2B-en\cite{uni31} and COYO-700M\cite{uni4}, and employs a pretrained ViT-L/14 CLIP backbone\cite{uni23-clip} for powerful multi-modal fusion. We chose VD for the final image reconstruction because it can integrate structural and semantic features during denoising.

\subsection{Visual Reconstruction from fMRI}
Decoding visual stimuli from brain activity remains a key challenge in computational neuroscience. Early approaches decoded visual stimuli from brain activity by extracting image features (e.g., multi-scale local bases \cite{11-3-08}, Gabor filters \cite{11-2}) and training linear mappings from fMRI signals to these features, demonstrating feasibility. However, fMRI’s low signal-to-noise ratio and limited datasets made natural scene reconstruction infeasible with linear methods.

Deep neural networks (DNNs) later enabled modeling complex brain-visual relationships through feature learning \cite{11-8}. Diverse DNN frameworks improved reconstruction, including deep belief networks \cite{10-4}, VAEs \cite{10-5-effenet}, feedforward architectures \cite{10-6,10-7}, GANs \cite{10-8,4-9}, and hybrid VAE/GANs \cite{7-12}, but often lacked semantic fidelity. Integrating CLIP \cite{uni23-clip} addressed this, Lin et al. \cite{2-7} aligned fMRI patterns with CLIP’s latent space via contrastive-adversarial learning, leveraging StyleGAN2 for semantically enhanced generation.

Recently, diffusion models has shown its ability to create high-resolution images with strong semantic consistency, leading to its successful use in various generative tasks\cite{3-27,3-28,3-29,3-7}. Takagi et al.\cite{9-1} was the first to map fMRI signals to diffusion latent space and CLIP text embeddings, generating images without training or fine-tuning deep networks, but the reconstructed images lacked sufficient semantic information and natural qualities. Later, MindEye\cite{12-4}, which optimized the semantic representations of fMRI features through contrastive learning, and MindDiffuser\cite{14-6}, which devised a two-stage diffusion process, addressed this issue. Furthermore, MindBridge\cite{13-5} introduces a cross-subject alignment strategy that enables the decoder to generalize across brain activity data from multiple individuals. 

Despite these advancements, existing methods still have limited capability in reconstructing visual stimuli of highly complex scenes, struggling to accurately capture their low-level structural features and high-level semantic information. This limitation hinders the generation of images that simultaneously achieve structural fidelity and semantic accuracy. 
\begin{figure*}[h] 
\centering 
\includegraphics[width=0.8\textwidth]{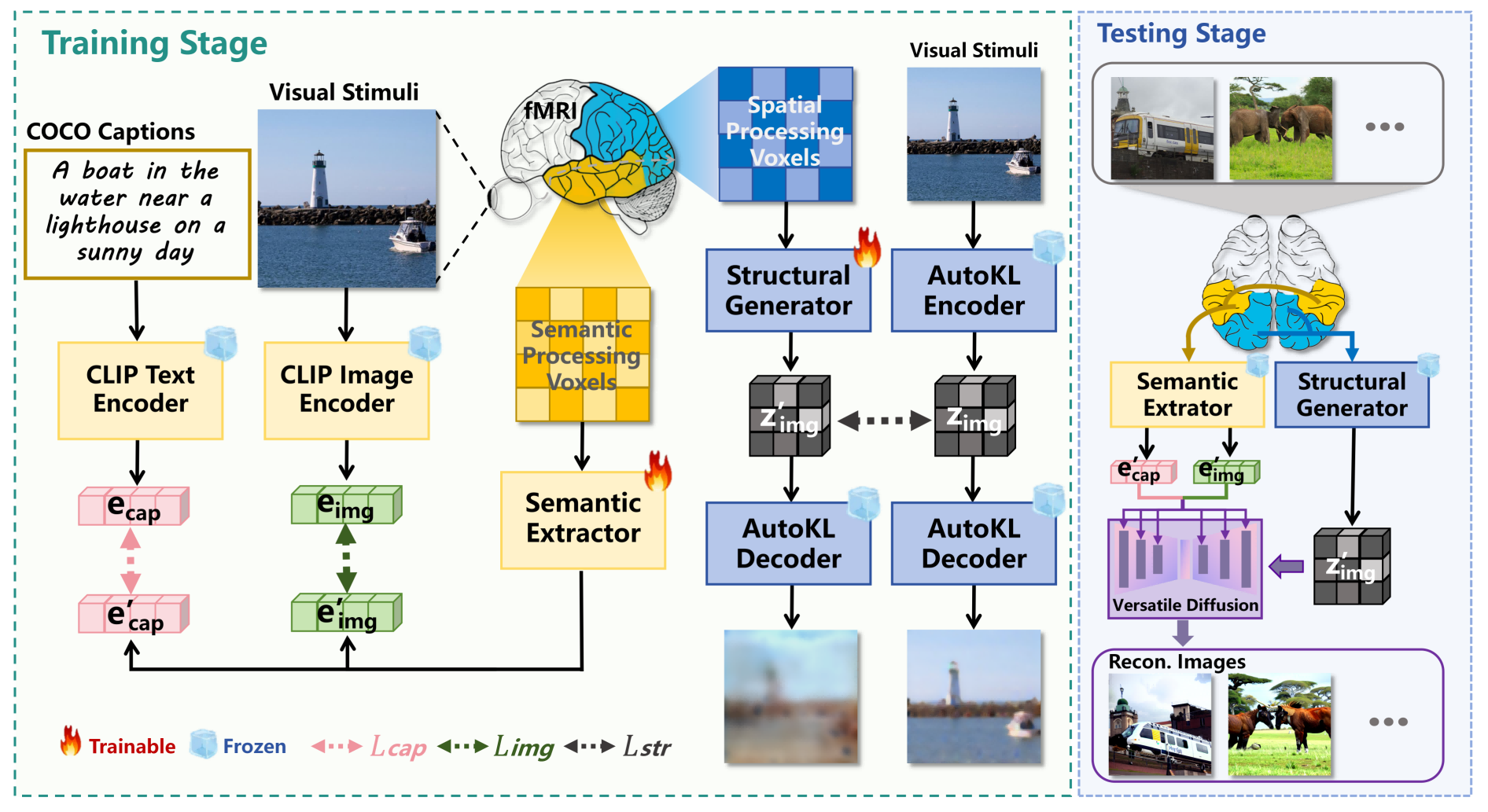}
\caption{
    Overall framework of HAVIR. Different data are used for training and testing.
}
\label{fig:1}
\end{figure*}

\section{Methods}
Our proposed framework HAVIR is composed of three components: Structural Generator, Semantic Extractor and a pre-trained Versatile Diffusion model. The framework of HAVIR is shown in Figure~\ref{fig:1}. 

\subsection{Structural Generator}
The Structural Generator converts spatial processing voxels into latent diffusion variables. Then the frozen pre-trained AutoKL Decoder, whose role in Versatile Diffusion is to decode the latent representation into pixel space, restores it as a predicted image. We call it Structural Generator because it captures fundamental spatial structures and color distributions that serve as a latent diffusion prior. However, it lacks detailed semantic content. See Appendix~\ref{stru gen} for components of the Structural Generator.

\textbf{Training objectives.} The training objective integrates two components: the \emph{MSE Loss} and the \emph{Sobel Loss}. These jointly ensure structural alignment and edge aware consistency between the predicted latent variable $\mathrm{z}'_{\text{img}}$ and the ground-truth latent variable $\mathrm{z}_{\text{img}}$, which is encoded from the visual stimulus using the AutoKL Encoder. 

\emph{MSE Loss} minimizes the mean squared error between the $\mathrm{z}'_{\text{img}}$ and the $\mathrm{z}_{\text{img}}$:
\begin{equation}  
    \mathcal{L}_{\text{mse}} = \frac{1}{n} \sum_{i=1}^{n} \left( \mathrm{z'}^{(i)}_{\text{img}} - \mathrm{z}^{(i)}_{\text{img}} \right)^2
\end{equation}
where \(n\) is the batch size. 

\emph{Sobel Loss} preserves high-frequency edge details through gradient magnitude align-ment\cite{sobel}, where the Sobel operator \(\mathcal{S}(\cdot)\) computes image gradients using horizontal (\(\mathbf{K}_x\)) and vertical (\(\mathbf{K}_y\)) convolution kernels (with \(\mathbf{K}_y = \mathbf{K}_x^\top\)):
\begin{equation}
\mathcal{L}_{\textit{sobel}} = \frac{1}{n} \sum_{i=1}^n \| \mathcal{S}(\mathrm{z'}_{\textit{img}}^{(i)}) - \mathcal{S}(\mathrm{z}_{\textit{img}}^{(i)}) \|_2^2
\end{equation}
The composite training loss is formulated by combining the above two objectives:
\begin{equation}  
    \mathcal{L}_{\text{str}} = \mathcal{L}_{\text{mse}} + \mathcal{L}_{\text{sobel}}
\end{equation}

\subsection{Semantic Extractor}
The Semantic Extractor is designed and trained to transform semantic processing voxels into CLIP image and text embeddings, which serve as conditional constraints for the denoising process. The term Semantic Extractor reflects its focus on capturing essential semantic information rather than structural features. See Appendix~\ref{seman extra} for components of the Semantic Extractor.

 \textbf{Training objectives.} The Semantic Extractor is trained to match its predicted embeddings to those generated by pretrained CLIP encoders. In particular, the predicted image embedding $\mathrm{e}'_{\text{img}}$ is aligned with the reference embedding $\mathrm{e}_{\text{img}}$ produced by a CLIP image encoder for each visual stimulus, and the predicted text embedding $\mathrm{e}'_{\text{cap}}$ is aligned with the reference embedding $\mathrm{e}_{\text{cap}}$ obtained from its corresponding caption via a CLIP text encoder. Captions are retrieved using the COCO ID associated with each stimulus.
 
 To achieve this, the Semantic Extractor framework employs two types of loss functions to optimize the mapping of CLIP image and text embeddings. Among them, the \textit{SoftCLIP Loss}\cite{12-4,13-5} has demonstrated effectiveness in aligning the fMRI modality with the pretrained CLIP embedding space. The function leverages a contrastive learning mechanism to maximize the similarity of positive pairs while minimizing the similarity of negative pairs:
\begin{equation}
\small
\begin{aligned}
\mathcal{L}_{SoftCLIP}(p, t) = - \sum_{i=1}^N \sum_{j=1}^N  
\Bigg[ &\frac{\exp(t_i \cdot t_j / \tau)}
{\sum_{m=1}^N \exp\left( \frac{t_i \cdot t_m}{\tau} \right)} &\\\times\log\left( \frac{\exp(p_i \cdot t_j / \tau)}
{\sum_{m=1}^N \exp\left( \frac{p_i \cdot t_m}{\tau} \right)} \right) \Bigg]
\end{aligned}
\end{equation}
Where \textit{p}, \textit{t} are the predicted and target CLIP embedding in a batch of size \textit{N}, respectively. $\tau$ is a temperature hyperparameter. However, using only the \textit{SoftCLIP Loss} causes noticeable artifacts because batch-wise soft labels create random variations, particularly with sparse and noisy fMRI-CLIP mappings. To address this, we introduced \textit{MSE Loss} to ensure direct consistency between predicted and target CLIP embeddings. The complete set of losses for predicting image and text CLIP embeddings incorporating these two losses:
\begin{equation}
\mathcal{L}_{\text{img}} = \mathcal{L}_{SoftCLIP}( \mathrm{e}_{\scriptstyle{\text{img}}} , \mathrm{e'}_{\scriptstyle{\text{img}}} ) + \mathcal{L}_{\text{mse}}( \mathrm{e}_{\scriptstyle{\text{img}}} , \mathrm{e'}_{\scriptstyle{\text{img}}} )
\end{equation}
\begin{equation}
\mathcal{L}_{\text{cap}} = \mathcal{L}_{SoftCLIP}( \mathrm{e}_{\scriptstyle{\text{cap}}} , \mathrm{e'}_{\scriptstyle{\text{cap}}} ) + \mathcal{L}_{\text{mse}}( \mathrm{e}_{\scriptstyle{\text{cap}}} , \mathrm{e'}_{\scriptstyle{\text{cap}}} )
\end{equation}

\subsection{Diffusion Reconstruction}
The diffusion process synthesizes images that preserve structural features while maintaining semantic information. 

\textbf{Noise controlled initialization.} The process initializes from the latent diffusion prior $\mathrm{z}'_{\text{img}}$ and adds partial noise controlled by the structural strength coefficient $\mathit{s} \in (0,1]$. The initial noise step $\tau = N - \lfloor N \cdot s \rfloor \quad$($N$ is the total timesteps) determines the noise level for denoising. The noised latent variable $\mathrm{z}_{\tau}$ is derived in one step:  
\begin{equation}
\mathrm{z}_{\tau} = \sqrt{\alpha_{\tau}} \mathrm{z’}_{\text{img}} + \sqrt{1 - \alpha_{\tau}} \epsilon, \quad \epsilon \sim \mathcal{N}(0, \mathit{I})
\end{equation}
where $\alpha_{\tau}$ is defined by the noise scheduler, $\epsilon$ is stochastic noise sampled from a standard normal distribution. 
\begin{figure*}[h] 
\centering 
\includegraphics[width=0.7\textwidth]{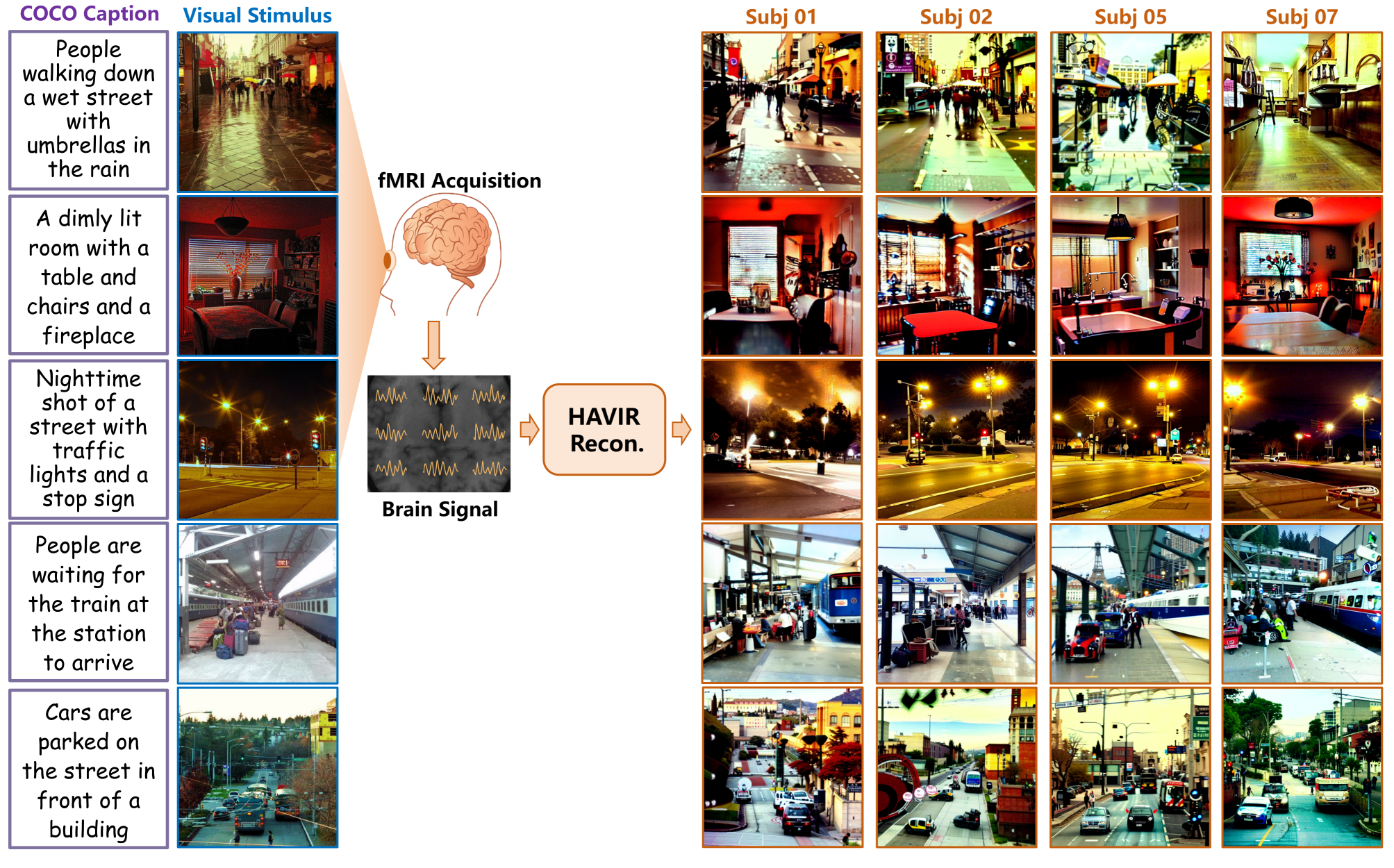}
\caption{
    Examples of the reconstruction results from HAVIR. 
}
\label{fig:2}
\end{figure*}

\textbf{Semantic conditioning.} The semantic constraints are imposed via cross-attention mechanisms, which guides the UNet during denoising. At each timestep $\tau_t$, the UNet predicts noise:  
\begin{equation}
\hat{\epsilon}_{\theta}(\mathrm{z}_t, \tau_t, \mathrm{e}_{\text{input}}) = 
\operatorname{UNet} \left(\mathrm{z}_t, \tau_t, 
\operatorname{CrosAtt}(\mathrm{e}'_{\text{img}}, \mathrm{e}'_{\text{cap}} )\right)
\end{equation}

\textbf{Guided denoising.}  
The denoising process iteratively refines the latent vector $\mathrm{z}_{\tau}$ through a sequence of timesteps $ t \in [\tau, 0] $. At each step $t$, the predicted noise $\hat{\epsilon}_{\theta}$ is used to update the latent state:
\begin{equation}
\mathrm{z}_{t-1} = 
\frac{1}{\sqrt{\alpha_t}} \left( 
\mathrm{z}_t - \frac{\beta_t}{\sqrt{1-\bar{\alpha}_t}} \hat{\epsilon}_\theta(\mathrm{z}_t, t, \mathrm{e}_{\text{input}})
\right)+ \sqrt{\beta_t} \cdot \boldsymbol{\epsilon}
\end{equation}
where $\beta_{t}$ is defined by the noise scheduler, $\alpha_t = 1 - \beta_t$, $\bar{\alpha}_t$ is the cumulative product of signal retention coefficients. This update rule progressively removes noise while integrating semantic information guided by the multimodal condition $\mathrm{e}_{\scriptstyle{\text{input}}}$.

\textbf{Image reconstruction.}
Finally, $\mathrm{z}_{\text{final}}$ is decoded into pixel space via the AutoKL Decoder $\mathcal{D}$, restoring the reconstructed image:  
\begin{equation}
\mathit{I}_{\scriptstyle{\text{recon}}} = \mathcal{D}( \mathrm{z}_{\scriptstyle{\text{final}}} )
\end{equation}
This diffusion process achieves structure-semantics decoupling: $\mathrm{z}'_{\text{img}}$ provides blurred low-level visual primitives, while $\mathrm{e}_{\text{input}}$ enforces high-level semantic concepts.

\section{Experiments}
\subsection{Data Elaboration}
Our study utilized the publicly available Natural Scenes Dataset (NSD)\cite{8-14}, a comprehensive 7T fMRI dataset collected from the brain signals of eight subjects during their viewing the COCO image \cite{imagenet-alex}. The corresponding textual caption for each image can be systematically retrieved through their associated COCO identifiers. Following previous neural decoding methodologies\cite{3-8,11-3-23,10-2,12-4,9-1,mdb47}, we analyzed data from four subjects (Subj01, 02, 05, 07) who completed all scanning sessions. Our experimental design includes: (a) a shared test set of the same 982 images shown to every subject, (b) different training sets containing 8,859 distinct images for each subject. This approach allows reliable evaluation of the model's performance using the same test set, while keeping adequate data for effective training.
\begin{table}[h]
\centering
\begin{minipage}[t]{0.517\linewidth}
  \includegraphics[width=\linewidth]{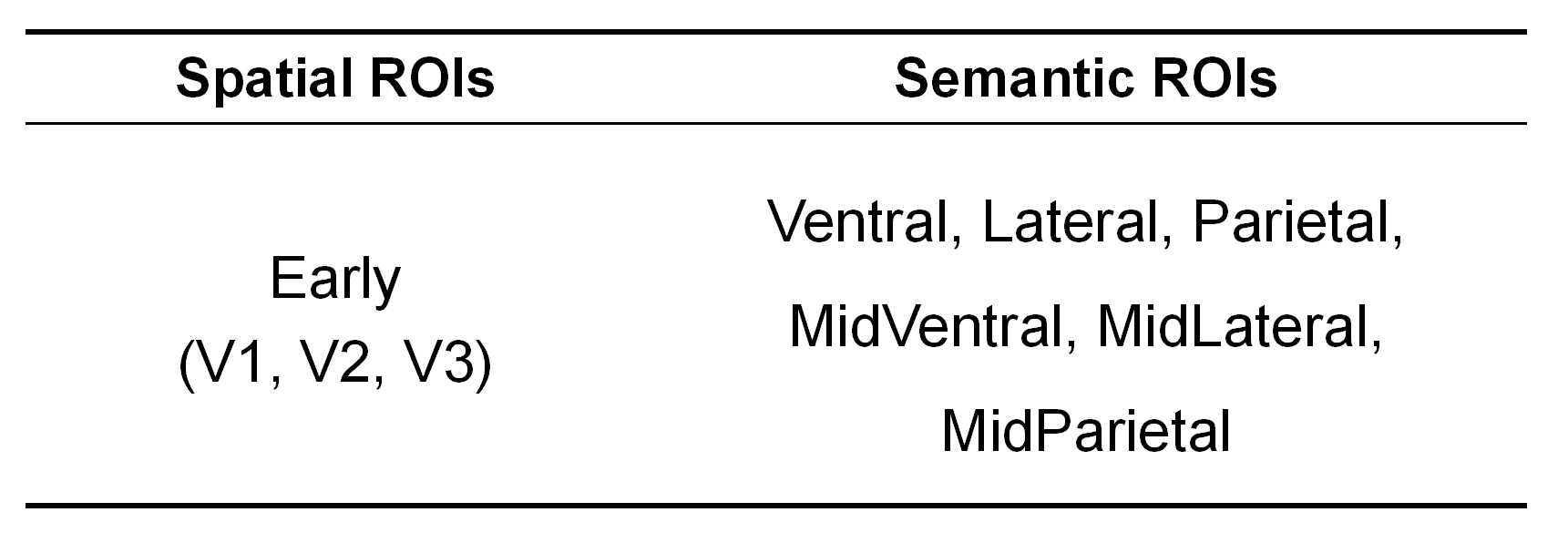}
  \caption{The details of the ROIs.}
  \label{tab:1}
\end{minipage}%
\hfill 
\begin{minipage}[t]{0.483\linewidth}
  \includegraphics[width=\linewidth]{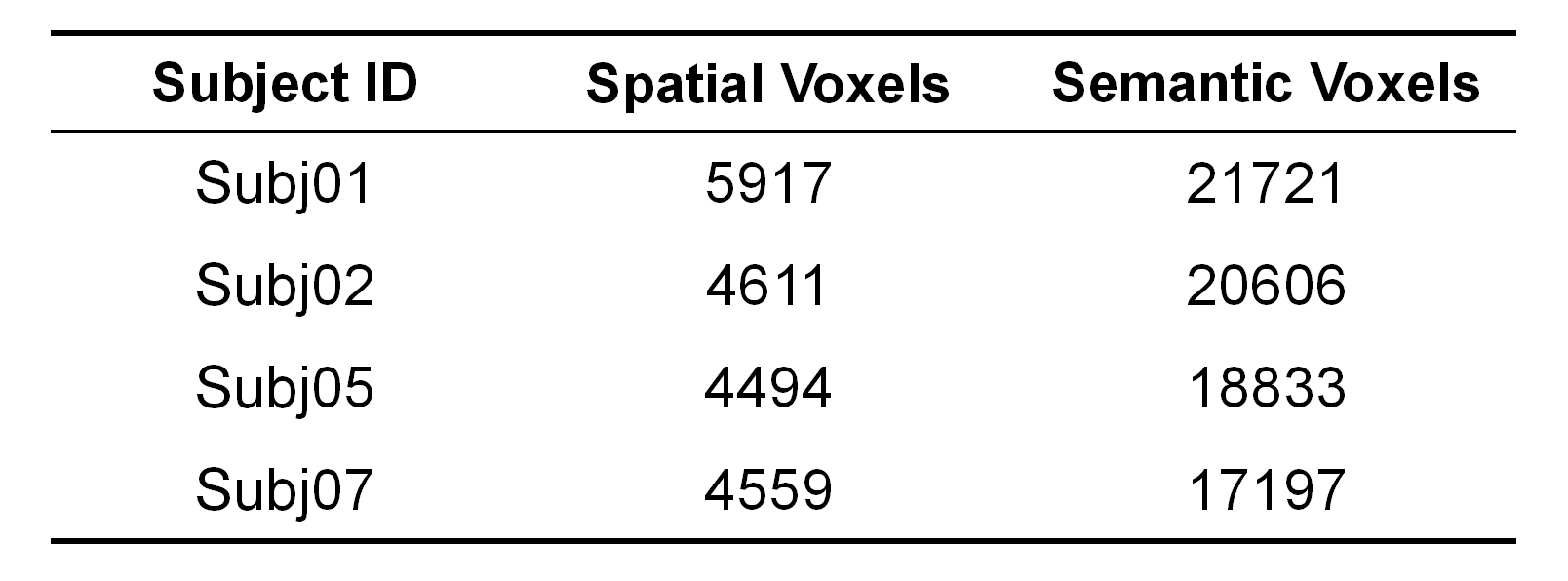}
  \caption{The voxel counts for each subject.}
  \label{tab:2}
\end{minipage}
\end{table}

We used preprocessed fMRI data with a spatial resolution of 1.8 mm. Using brain region masks provided by the NSD, we applied boolean indexing to extract voxels from the visual cortex. These extracted voxels were then categorized into spatial and semantic subsets (see Table~\ref{tab:1} for ROI definitions). Table~\ref{tab:2} quantifies the distribution of voxels for each subject within these ROIs. 

\textbf{Notably}, previous studies have primarily relied on standardized brain templates for spatial parcellation. However, both neuroanatomical architecture and functional organization demonstrate significant inter-individual variability\cite{visualdiff}. During spatial normalization, these inter-subject discrepancies can induce boundary blurring or registration inaccuracies, ultimately compromising model training efficacy and image reconstruction performance. In contrast to conventional approaches, our study utilizes NSD provided brain masks incorporating manually delineated ROI boundaries for individual subjects. These customized masks account for neuroanatomical and functional specificity through fine grained regional mapping, enabling differentiation of areas with subtle functional distinctions, thereby enhancing downstream processing precision.

\subsection{Visual Reconstruction Examples}
To demonstrate our method's capability in reconstructing both semantic details and spatial structures of complex scenes, we selected five high complexity visual stimuli for analysis. As shown in Figure~\ref{fig:2}, the six-column layout is organized as follows: the first column displays COCO captions, the second shows visual stimuli (ground truth), while the right four columns present individualized reconstruction results from four subjects.
\begin{figure*} 
\centering 
\includegraphics[width=0.95\textwidth]{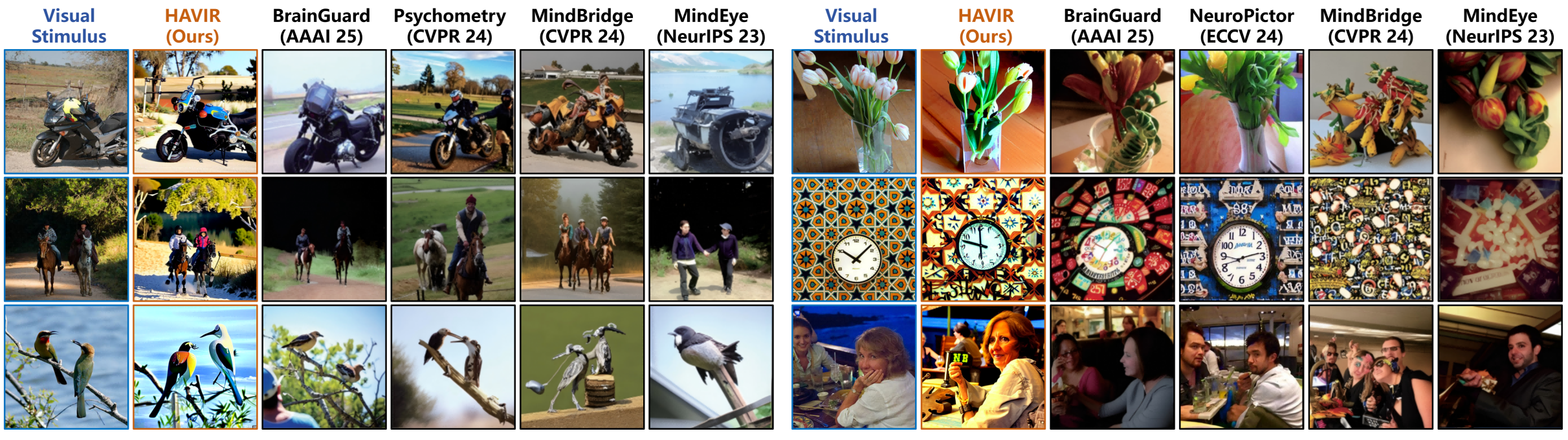}
\caption{
    Qualitative comparisons on the NSD test dataset. The results of HAVIR demonstrate superior reconstruction accuracy compared to the five recent SOTA methods. 
}
\label{fig:3}
\end{figure*}

The results indicate our model effectively reconstructs both the spatial layouts and semantic information of complex scenes. For example, the first three visual stimuli include a rain-soaked sidewalk with pale yellow background, a kitchen scene under dark red lighting, and a nighttime street with flickering streetlights. The reconstructions from all four subjects not only accurately preserve spatial structures but also precisely reproduce essential visual characteristics, such as the pale yellow ambient light, the dark red light distribution in the kitchen, and the dynamic flickering of streetlights.

\subsection{Comparison to Existing Methods}
In this section, we compare HAVIR with five state-of-the-art(SOTA) methods: BrainGuard\cite{BrainGuard}, NeuroPictor\cite{Neuropictor}, Psychometry\cite{psychometry}, MindBridge\cite{13-5}, and MindEye\cite{12-4}.

\textbf{Qualitative results.} For qualitative evaluation, our reconstructed images are visually compared with their results, as shown in Figure~\ref{fig:3}. Whereas existing approaches exhibit reduced performance on complex scenes, HAVIR maintains high reconstruction fidelity under such challenging conditions. This capability is exemplified by the two cases shown on the right side of Figure~\ref{fig:3}. In the first case, comparative approaches failed to reproduce the floral color accurately, whereas HAVIR successfully captured the key semantic feature of pink flowers. In the second case, our method simultaneously recovered background textures and the clock's spatial position, a combination that none of the comparative methods managed to achieve.

\textbf{Quantitative evaluation.} To objectively evaluate the reconstruction performance, we conduct quantitative evaluation using the following metrics. At the low-level evaluation, the Structural Similarity Index (\textbf{SSIM}) quantifies spatial coherence and perceptual integrity \cite{pixcor-ssim}, complemented by Pixel Correlation (\textbf{PixCorr}) for evaluating pixel-wise accuracy \cite{pixcor-ssim}. Mid-level analysis utilizes \textbf{AlexNet(2/5)} to measure texture pattern consistency through feature similarity comparisons \cite{imagenet-alex}. For high-level semantic evaluation, Inception score (\textbf{Incep}) evaluates the image quality and diversity \cite{inception}, \textbf{CLIP} evaluates vision-language embedding consistency \cite{uni23-clip}, \textbf{EffNet-B} computes object-discriminative feature distances \cite{10-5-effenet}, and \textbf{SwAV} quantifies semantic divergence in contrastive learning space \cite{swav}. For PixCorr, SSIM, Alex(2), Alex(5), Incep, and CLIP metrics, higher is the better. For EffNet-B and SwAV distances, lower is the better.
\begin{table}[h]
\centering
\includegraphics[width=\linewidth]{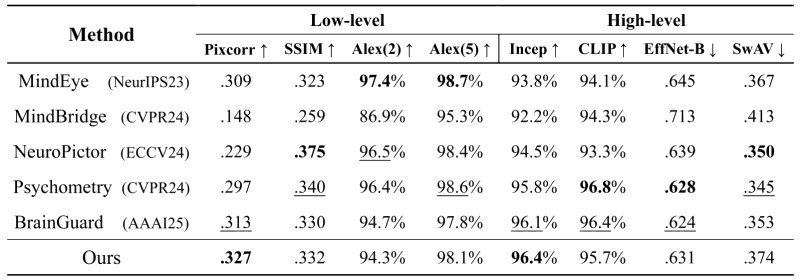}
\caption{Quantitative comparison with SOTA methods. Bold indicates the best, underline denotes the second-best.}
\label{tab:3}
\end{table}

As shown in Table~\ref{tab:3}, HAVIR demonstrates superior performance across multiple levels in the quantitative evaluation. It achieves the highest PixCorr (0.327) and ranks third in SSIM (0.332) among low-level metrics, demonstrating strong performance in both pixel-level accuracy and structural preservation. In mid-level feature analysis, the AlexNet(5) achieves a near-optimal score of 98.1\%, indicating its strong capacity to capture global texture patterns. These results are highly consistent with qualitative findings, such as the reconstruction of clock spatial position and background texture, suggesting precise detail control in the reconstruction of complex scenes. In high-level semantic evaluation, HAVIR tops the charts with an Inception Score of 96.4\%, further confirming its leading performance in semantic fidelity (e.g., accurately recovering key attributes such as flower color).

\subsection{Ablation Studies}
In the ablation studies, we evaluate the contribution of each component in HAVIR using four modified configurations: \textbf{only $\mathrm{z}'_{\text{img}}$} retains only structural pipeline's latent prior without CLIP guidance, \textbf{w/o $\mathrm{e}'_{\text{cap}}$} removes text embedding guidance, \textbf{w/o $\mathrm{z}'_{\text{img}}$} removes image embedding guidance, and \textbf{w/o $\mathrm{z}'_{\text{img}}$} replaces structural prior with Gaussian noise initialization. This systematic comparison evaluates the individual impacts of structural priors, text semantics, and image features on reconstruction quality.
\begin{figure} 
\centering 
\includegraphics[width=0.5\textwidth]{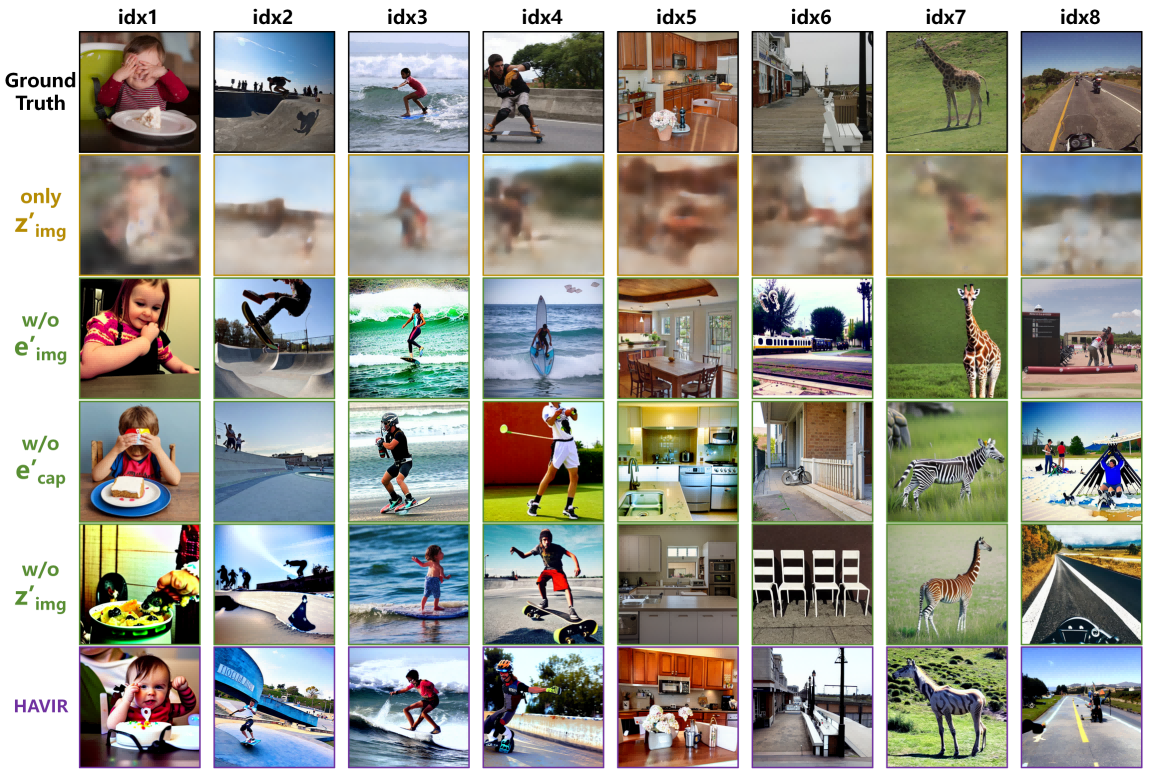}
\caption{
    Qualitative results of the full model and its ablated configurations 
}
\label{fig:4}
\end{figure}

\textbf{Qualitative results.}
As shown in Figure~\ref{fig:4}, the qualitative results reveal the functions of each component. In the ``only $\mathrm{z}'_{\text{img}}$" configuration, the reconstructions preserve blurry object layouts, but the absence of CLIP guidance leads to a complete loss of semantic information. The ``w/o $\mathrm{e}'_{\text{img}}$" and ``w/o $\mathrm{e}'_{\text{cap}}$" configurations can restore the spatial structure of the ground truth, but show deviations in semantic accuracy (e.g., idx4's skating on the road and idx8's cycling scene are not correctly reconstructed). In contrast, the ``w/o $\mathrm{z}'_{\text{img}}$" configuration shows the opposite effect: while semantic information is accurately reconstructed (e.g., the core semantic content of the chair in idx6), the spatial structure deviates systematically from the ground truth. These experiments indicate that during the HAVIR reconstruction process, the structural prior ($\mathrm{z}'_{\text{img}}$) primarily governs the fidelity of spatial topology, while the dual-modal embeddings of CLIP ($\mathrm{e}'_{\text{cap}}$ and $\mathrm{e}'_{\text{img}}$) jointly constrain the accurate recovery of high-level semantics.
\begin{table}[h]
\centering
\includegraphics[width=\linewidth]{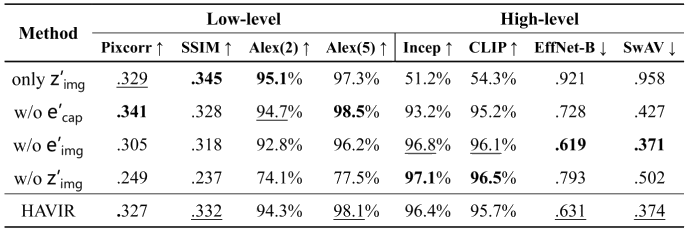}
\caption{Comparison of the full model with its ablated configurations in terms of quantitative results.}
\label{tab:4}
\end{table}

\textbf{Quantitative evaluation.} The quantitative results in Table~\ref{tab:4} further validate the qualitative findings. When relying solely on structural priors(only $\mathrm{z}'_{\text{img}}$), mid-level features (AlexNet(2/5)) and low-level metrics (Pixcorr, SSIM) remain acceptable, but high-level semantics collapse entirely (CLIP=54.3\%, SwAV=0.958). Conversely, removing structural priors (w/o $\mathrm{z}'_{\text{img}}$) causes a sharp decline in low-level metrics (SSIM=0.237), while dual-modal embeddings still preserve high-level semantics (CLIP=96.5\%). However, removing either modality CLIP embedding (w/o $\mathrm{e}'_{\text{cap}}$ or w/o $\mathrm{e}'_{\text{img}}$) achieves near-optimal high-level scores, while leading to semantic inaccuracy, reflected in degraded EffNet-B and SwAV metrics. Ultimately, the full model (HAVIR) achieves balanced optimization across all metrics by synergizing structural prior and semantic constraint, reflecting the necessity of multimodal joint guidance.
\begin{figure} 
\centering 
\includegraphics[width=0.5\textwidth]{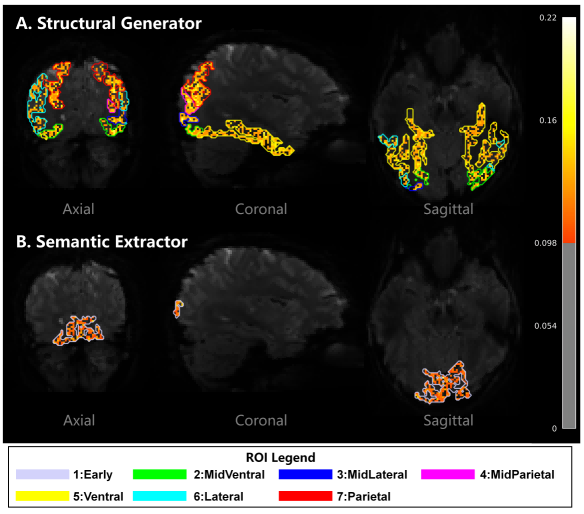}
\caption{
    Spatial mapping of brain region contributions to Structural Generator (A) and Semantic Extractor (B) on Subj01.
}
\label{fig:5}
\end{figure}
\begin{figure*}[h] 
\centering 
\includegraphics[width=\textwidth]{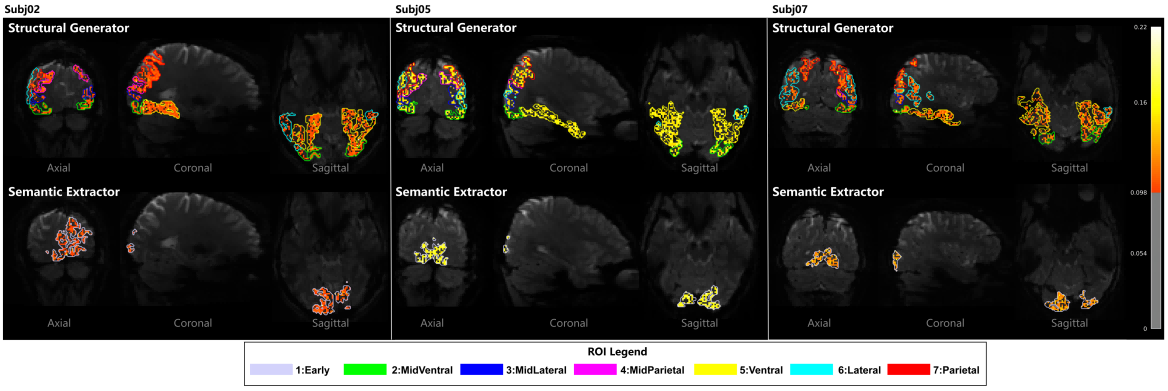}
\caption{
    Voxel-wise contribution intensity maps for Subj02, Subj05 and Subj07, demonstrating cross-subject adaptability.}
\label{fig:6}
\end{figure*}
\subsection{Interpretability Analysis of HAVIR}
To explore the decoding mechanisms underlying HAVIR's reconstruction of visual information and to evaluate its adaptability to cross-subject functional brain representations, we conduct explainability analyses by inversely mapping the model parameters onto functional connectivity patterns. Specifically, we extracted the weight matrices from the fully connected layers of the Structural Generator and Semantic Extractor, respectively. For each voxel, we calculated the L2 norm of its corresponding weight vector to quantify its contribution intensity to hidden layer activation. Subsequently, these values were projected back into the whole-brain space to generate a three-dimensional brain region importance distribution map. Figure~\ref{fig:5} shows the results on Subj01, with other subjects' results provided in Figure~\ref{fig:6}. 

The experimental results demonstrate the adaptability of HAVIR to cross-subject brain characteristics: despite notable anatomical differences between the four subjects, the model accurately matched each person's unique functional brain patterns by adjusting the weight across different brain regions. This indicates that the model does not simply apply a uniform template, but rather dynamically constructs decoding pathways tailored to each individual by analyzing their specific brain features. This personalized adaptation preserves the uniqueness of each subject's visual information processing and is a key factor enabling the model to stably reconstruct what different subjects see.

\section{Conclusion}
In this paper, we first discuss the limitations of existing methods in reconstructing complex scenes and then propose HAVIR, a model inspired by visual processing mechanisms to address this issue. HAVIR decomposes fMRI signals into structural and semantic processing voxels via two modules: a Structural Generator capturing spatial patterns and a Semantic Extractor decoding conceptual content. Experiments demonstrate that our approach surpasses stateof-the-art methods across multiple evaluation metrics.

\appendix
\subsection{Components of the Structural Generator}
\label{stru gen}
The Structural Generator first projects spatial processing voxels into a hidden space via linear mapping, followed by sequential processing through LayerNorm, SiLU, and Dropout. These representations are then transformed by residual MLP blocks before being passed to a second linear layer that generates low-resolution feature maps. The maps undergo GroupNorm normalization and upsampling to produce the output $\mathrm{z}'_{\text{img}}$, ensuring alignment with the latent vector $\mathrm{z}_{\text{img}}$, which is the variable encoded by the frozen AutoKL Encoder from visual stimulus.

\subsection{Components of the Semantic Extractor}
\label{seman extra}
The Semantic Extractor uses an embedding network to transform semantic processing voxels into a hidden space via linear projection, normalization, activation, and dropout, while a residual MLP refines features through cascaded blocks for stable gradients and semantic reinforcement. Subsequently separate linear heads project features into CLIP’s multimodal spaces: the image head to $\mathrm{e}'_{\text{img}}$ (CLIP’s image encoder) and the text head to $\mathrm{e}'_{\text{cap}}$ (CLIP’s text encoder), maintaining distinct semantics while preserving CLIP’s contrastive alignment. This architecture separates semantic enhancement and modality projection, maintaining compatibility with CLIP's pretrained space while enabling authentic semantic information derived from semantic processing voxels.

\subsection{Implementation Details}
The model was initially trained from scratch for 200 epochs using 40 hours of Subj01's fMRI data on a single NVIDIA A800 GPU (80GB) with a batch size of 30. This base model was subsequently fine-tuned through transfer learning using 1-hour datasets from Subj02, 05, and 07, performed across three NVIDIA RTX 4090 GPUs (24GB) with a cumulative batch size of 50.

\bibliography{egbib}
\end{document}